\documentclass[letterpaper]{article} 
\usepackage{aaai2026}  
\usepackage{times}  
\usepackage{helvet}  
\usepackage{courier}  
\usepackage[hyphens]{url}  
\usepackage{graphicx} 
\usepackage{natbib}  
\usepackage{caption} 
\usepackage{algorithm}
\usepackage{algorithmic}

\usepackage{newfloat}
\usepackage{listings}

\usepackage{subcaption}
\usepackage{xcolor}
\usepackage{colortbl}
\usepackage{multirow}
\usepackage{tikz}
\usepackage{overpic}
\usepackage{tablefootnote}
\usepackage{amsmath} 
\usepackage{arydshln}
\usepackage{makecell} 
\usepackage{array}
\usepackage{color}
\usepackage{enumitem}

\definecolor{LightCyan}{rgb}{0.88,1,1}

\DeclareCaptionStyle{ruled}{labelfont=normalfont,labelsep=colon,strut=off} 
\floatstyle{ruled}
\newfloat{listing}{tb}{lst}{}
\floatname{listing}{Listing}
\title{Realistic Face Reconstruction from Facial Embeddings via Diffusion Models}

\author {
    Dong Han\textsuperscript{\rm 1,\rm 2}, Yong Li\textsuperscript{\rm 1}\thanks{Corresponding author.}, Joachim Denzler\textsuperscript{\rm 2} 
}
\affiliations {
    \textsuperscript{\rm 1}Data Protection Technology Lab, Huawei Technologies Düsseldorf, Germany\\
    \textsuperscript{\rm 2}Computer Vision Group, Friedrich Schiller University Jena, Germany\\
    \{dong.han2, yong.li1\}@huawei.com, \{dong.han, joachim.denzler\}@uni-jena.de
}

\begin{document}

\maketitle

\begin{abstract}

With the advancement of face recognition (FR) systems, privacy-preserving face recognition (PPFR) systems have gained popularity for their accurate recognition, enhanced facial privacy protection, and robustness to various attacks.
However, there are limited studies to further verify privacy risks by 
reconstructing realistic high-resolution face images from embeddings of these systems, especially for PPFR.
In this work, 
we propose the face embedding mapping (FEM), a general framework 
that explores Kolmogorov-Arnold Network (KAN) for conducting the embedding-to-face attack by leveraging pre-trained Identity-Preserving diffusion model against state-of-the-art (SOTA) FR and PPFR systems.
Based on extensive experiments, we verify that reconstructed faces can be used for accessing other real-word FR systems.
Besides, the proposed method shows the robustness in reconstructing faces from the partial and protected face embeddings.
Moreover, FEM can be utilized as a tool for evaluating safety of FR and PPFR systems in terms of privacy leakage. All images used in this work are from public datasets.

\end{abstract}

\section{Introduction}
\label{sec:intro}

The advancement of artificial intelligence has brought attention to the security and privacy concerns associated with biometric authentication systems \cite{laishram2024toward,wang2024make},
specifically for face recognition (FR) \cite{rezgui2024enhancing}.
FR systems generate the template for each identity for comparing different faces and to authenticate query faces. These face templates or face embeddings
are considered as one type of biometric data that is frequently produced by \textit{black-box} models (e.g., convolutional neural networks (CNNs) and deep neural networks (DNNs) models).
\begin{figure}[t!]
    \centering
    \includegraphics[scale=0.36]{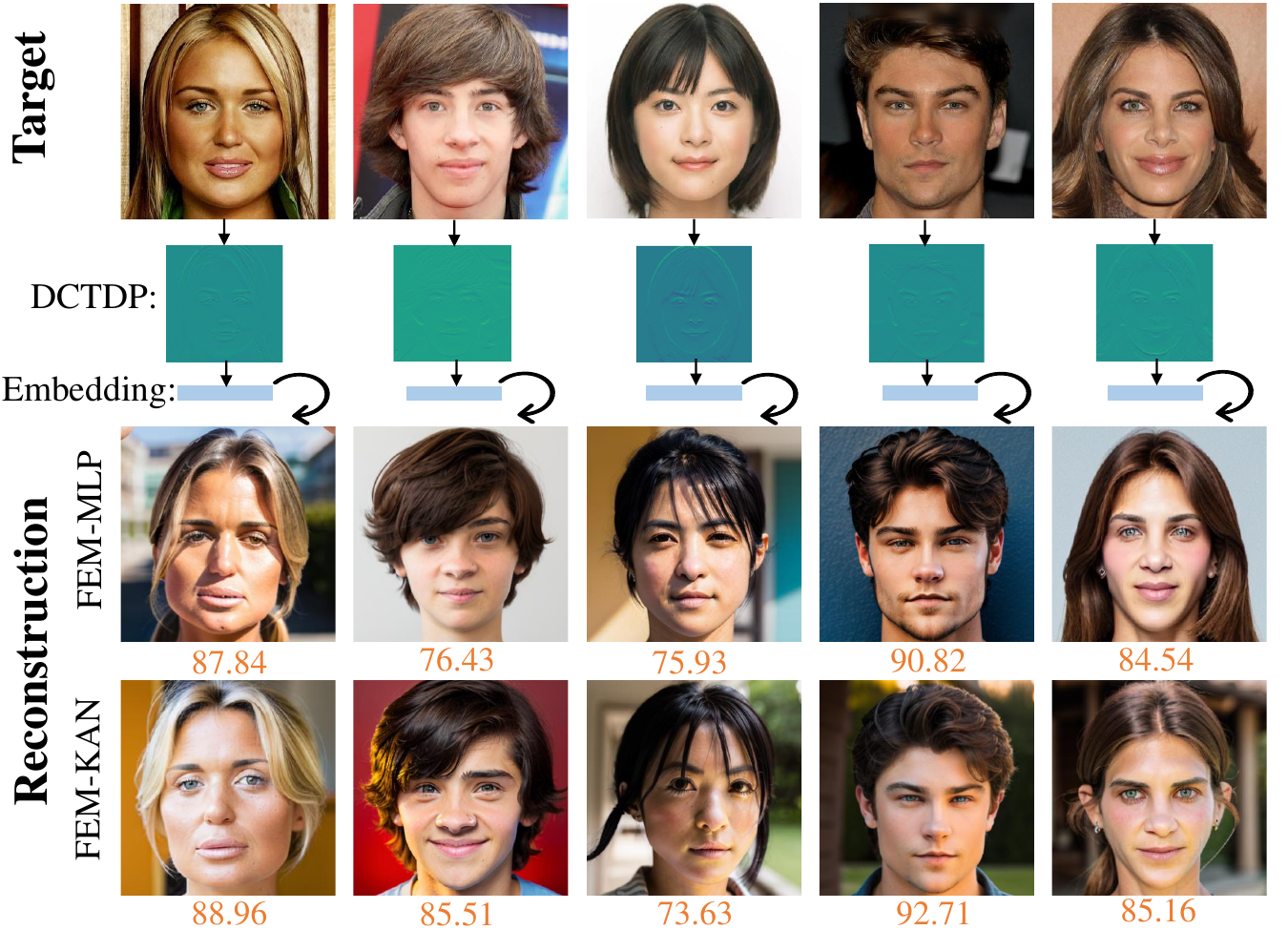}
    \caption{Sample face images from the CelebA-HQ dataset (first
    row) and their corresponding reconstructed face images from face embeddings of PPFR model DCTDP. The \textcolor{orange}
    {orange color value} indicates confidence score (higher is better) given by commercial API Face++.}
    \label{fig:reconstructed faces}
\end{figure}
Existing common threats to the face embeddings are sensitive information retrieval attacks \cite{terhorst2020beyond} (extract soft-biometric information such as sex, age, race, etc.)
or face reconstruction attacks (recover the complete face image from an embedding).
To increase the privacy and security level of FR, privacy-preserving face recognition (PPFR) systems \cite{DCTDP, PartialFace, HFCF, MinusFace} have been proposed.
However, most PPFR methods only focus on concealing visual information from input face images to systems while embeddings are not being protected directly.
To protect face embeddings, various protection algorithms are proposed.
Hash-based method \cite{IoM} transfers embeddings into discrete index hashed code.
Homomorphic Encryption (HE)-based method \cite{homomorphic_encryption} encrypts embedding into ciphertext.
However, these methods require high computational cost and degrade recognition.
Recently, transform-based methods offers more efficient protection.
PolyProtect \cite{PolyProtect} is based on multivariate polynomials with user-specific parameters to map embeddings.
MLP-Hash \cite{Mlp-hash} utilizes user-specific randomly-weighted multi-layer perceptron (MLP) to rotate and binarize embeddings.
SlerpFace \cite{slerpface} employs spherical linear interpolation to rotate embedding towards noise.

\begin{figure*}[ht!]
    \centering
    \includegraphics[scale=0.42]{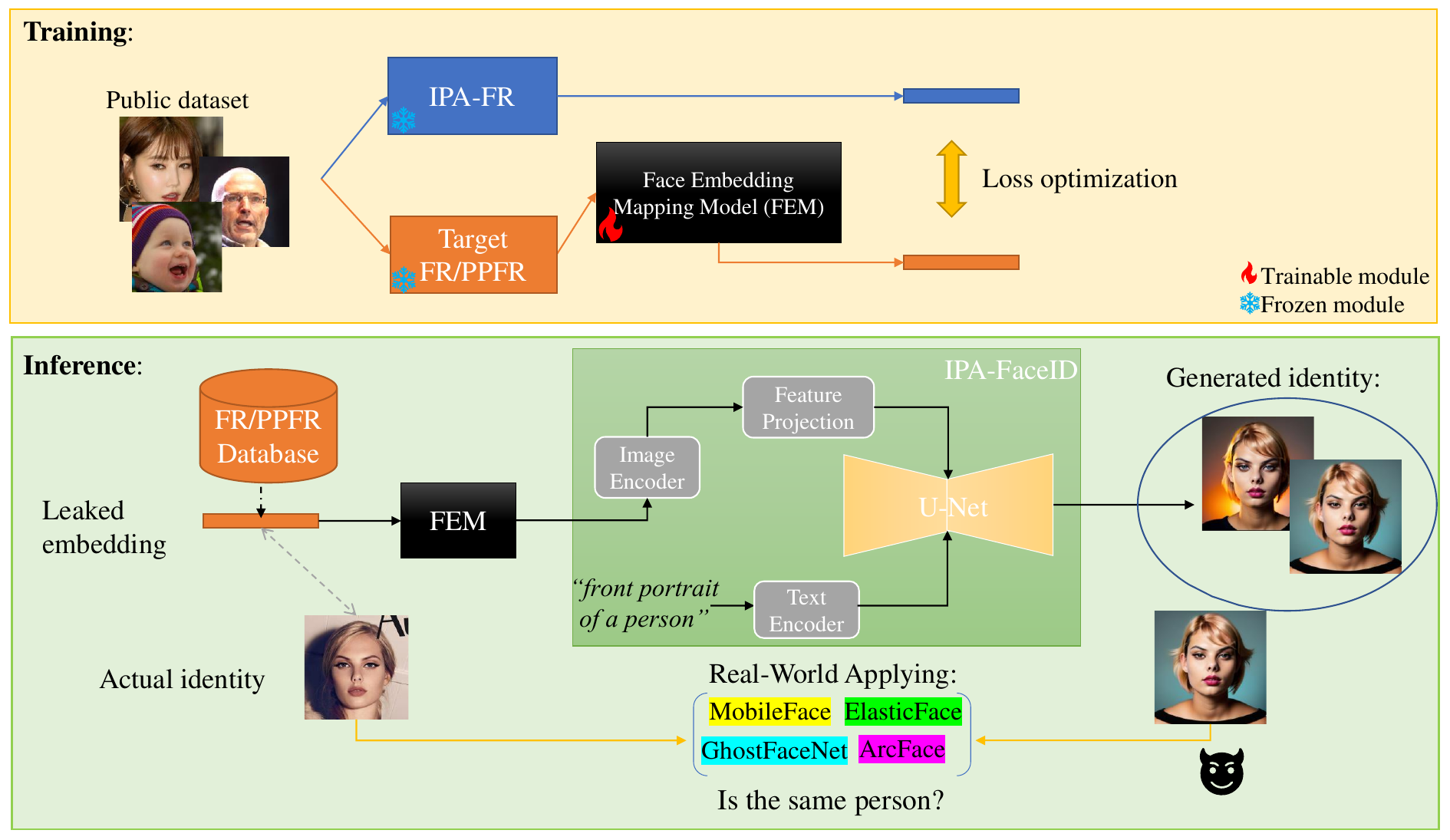}
    \caption{Pipeline of face embedding mapping (FEM). In training, FEM learns to map between target model and the default FR of IPA-FaceID. During inference stage, trained FEM can directly reconstruct complete face images from the leaked embedding.}
    \label{fig:face reconstruction}
\end{figure*}

Current face reconstruction methods focus on face image reconstruction from embeddings of normal FR models (without designed operations for privacy protection on either input face images or face embeddings).
Deconvolutional neural network NbNet \cite{NbNet} utilizes the deconvolution to reconstruct face images from deep templates.
End-to-end CNN-based method \cite{CNN_FACE_Recover} combines cascaded convolutional layers and deconvolutional layers to improve reconstruction.
Moreover, the learning-based method \cite{shahreza2024breaking} can reconstruct the underlying face image from a protected embedding that is protected by embedding protection mechanisms \cite{Biohashing, Mlp-hash, homomorphic_encryption}.
Nevertheless, the reconstructed faces from these methods suffer from noisy and blurry artifacts, which degrade the image naturalness.
Generative adversarial network (GAN)-based approach FaceTI \cite{MappingNetworks} trains a mapping network to transfer face embedding to the latent space of a pre-trained face generation network.
However, they only test their method on normal FR systems and the whole training process is resource-intensive.
Recently, MAP2V explores reconstructing face images from PPFR systems by employing GAN model without training.   
Nevertheless, it requires long inference time which limits its potential in real-time application.  
\citeauthor{shahreza2025face} leverage a pre-trained face foundation model \cite{arc2face} for face reconstruction. However, it is only tested on normal FR systems. 
In our work, we choose FaceTI and MAP2V as the main comparisons due to their superiority in realistic face image reconstruction compared with previous CNN-based methods.
Considering the above motivations, 
we propose a face reconstruction framework to generate realistic face identity (ID) based on leaked face embeddings from both FR and PPFR models, by utilizing a pre-trained ID-Preserving diffusion model, IPA-FaceID \cite{ip-adapter}.
As depicted in Fig. \ref{fig:face reconstruction}, we feed training face images to both IPA-FR (default FR of IPA-FaceID) and target FR models.
The initial output face embedding from the target FR model is transferred by the Face Embedding Mapping (FEM) model
before performing loss optimization.
During the inference stage, the leaked embedding from the target FR or PPFR model can be mapped by trained FEM and directly used by IPA-FaceID to generate realistic face images.
We verify the effectiveness of face reconstruction by applying impersonation attacks to real-world FR systems.
Besides, reconstructed face images by FEMs can also bypass the commercial face comparison API, Face++\footnote{https://www.faceplusplus.com} as shown in Fig. \ref{fig:reconstructed faces}.

Our key contributions are:
\begin{itemize}
    \item We propose FEM: a face embedding mapping framework that maps the arbitrary embedding of target FR or PPFR for realistic face reconstruction by utilizing pre-trained ID-Preserving diffusion model.
    \item We exploit the potential of KAN for face embedding mapping and showcase the efficacy of the FEM-KAN model for non-linear mapping compared with SOTA face reconstruction models. 
    \item Through extensive experiments, we demonstrate that proposed method improves on face reconstruction performance and show the generalization to partial and protective embedding
\end{itemize}

\section{Related Work}

\subsection{Identity-Preserving T2I Diffusion Models}

Existing text-to-image (T2I) models still have limitations in generating accurate and realistic images due to the limited information expressed by text prompts. 
IP-Adapter \cite{ip-adapter} proposes decoupled cross-attention to embed image feature to a pre-trained T2I diffusion model by adding a new cross-attention layer. 
It utilizes a trainable projection model to map the image embedding that extracted by a pre-trained CLIP image encoder \cite{clipencoder} to a sequence image feature.
Later on, IPA-FaceID\footnote{https://huggingface.co/h94/IP-Adapter-FaceID} is developed for customized face image generation by
integrating face information through face embedding extracted from a FR model instead of CLIP image embedding.
IPA-FaceID has the ability to produce corresponding diverse styles of image based on a given face image and text prompt.
InstantID \cite{instantid} proposes a trainable lightweight module for transferring face features from the frozen face encoder into the same space of the text token.
It leverages modified ControlNet \cite{ControlNet} for identity preservation.
Arc2Face \cite{arc2face} is dedicated for ID-to-face generation by using only
ID embedding. It fixes the text prompt with a frozen pseudo-prompt ``a photo of $\left\langle id \right\rangle$ person''
where placeholder $\left\langle id \right\rangle$ token embedding is replaced by ArcFace embedding of image prompt. 

\subsection{From Deep Face Embeddings to Face images}

Extracting images from deep face embeddings is challenging for naive deep learning networks e.g., UNet \cite{UNet}.
CNN-based network \cite{CNN_FACE_Recover, shahreza2023blackbox} is introduced to reconstruct face images from corresponding embeddings that are extracted from the FR model by end-to-end training.
With more restrictions on the embedding leakage of FR models, previous work \cite{partial_face_reconstruction} attempts to
reconstruct the underlying face image from partial leaked face embeddings by using a similar face reconstruction network as \cite{CNN_FACE_Recover}.
However, reconstructed face images from these methods are highly blurred. 
Furthermore, DSCasConv (cascaded convolutions and skip connections) is proposed to reduce the blurring \cite{shahreza2024vulnerability}. 
However, it still has noticeable blurry artifacts around face contour.
Most of CNN-based face reconstruction methods focus on reconstructing low-resolution face images.
For more realistic face reconstruction, FaceTI \cite{MappingNetworks} takes advantage of GAN model to generate face images from the deep face embedding.
They employ the pre-trained StyleGAN3 \cite{StyleGAN3} network to establish a mapping from facial embeddings to the intermediate latent space of StyleGAN.
Recently, MAP2V \cite{zhang2024validating} is a training-free approach that explores reconstructing face images from PPFRs by utilizing StyleGAN to construct prior space and ranks the top-k latent vectors for 
reconstruction. \citeauthor{shahreza2025face} train an adapter (a linear layer) to map the face embedding and utilize a pre-trained face foundation model for reconstruction.

\section{Approach}

\subsection{Problem Formulation}

\textbf{Attacker's Goal.} Consider a target FR system, the attacker aims to reconstruct a complete face image from a face embedding.
Then the reconstructed face image is used to query the same or a different FR system.

\textbf{Attacker's Knowledge.} The attacker has access to a leaked face embedding of a user registered in the FR system's database.
The attacker only has \textit{black-box} knowledge of the
feature extractor model in the same FR system, i.e., can only obtain features from query image.
The attacker can inject the reconstructed face image into the target FR systems.

\textbf{Attacker's Strategy.} Though the attacker has no access to the private face dataset that the target model is being trained on,
attacker can collect a public dataset from the Internet or generate a synthetic face dataset for training a face reconstruction model to map face embeddings and reconstruct underlying face images.
Then, the attacker can utilize the reconstructed facial images to gain access to the target FR system.

\subsection{Kolmogorov-Arnold Theorem Preliminaries}

The Kolmogorov-Arnold theorem \cite{kan} states that any continuous function may be expressed as a combination of a finite number of continuous univariate functions. 
For every continuous function $f(x)$ defined in the n-dimensional real space, where $x = (x_1, x_2, ..., x_n)$, it can be represented as a combination of a univariate continuous function $\Phi$ and a sequence of continuous bivariate functions $x_{i}$ and $\phi_{q,i}$. 
The theorem demonstrates the existence of such a representation: $f(x) = \sum_{q} \Phi _{q}(\sum_{i}\phi _{q,i}(x_{i}))$.
It suggests that even sophisticated functions in high-dimensional spaces can be reconstructed through a sequence of lower-dimensional function operations.
The mapping between face embeddings from different systems can be exactly represented as a composition of univariate functions and additions. 
For face embeddings (typically high-dimensional but structured), this decomposition can better capture complex, non-linear relationships between embedding spaces.

\subsection{Face Embedding Mapping (FEM)}

\begin{figure}[t!]
    \centering
    \includegraphics[scale=0.38]{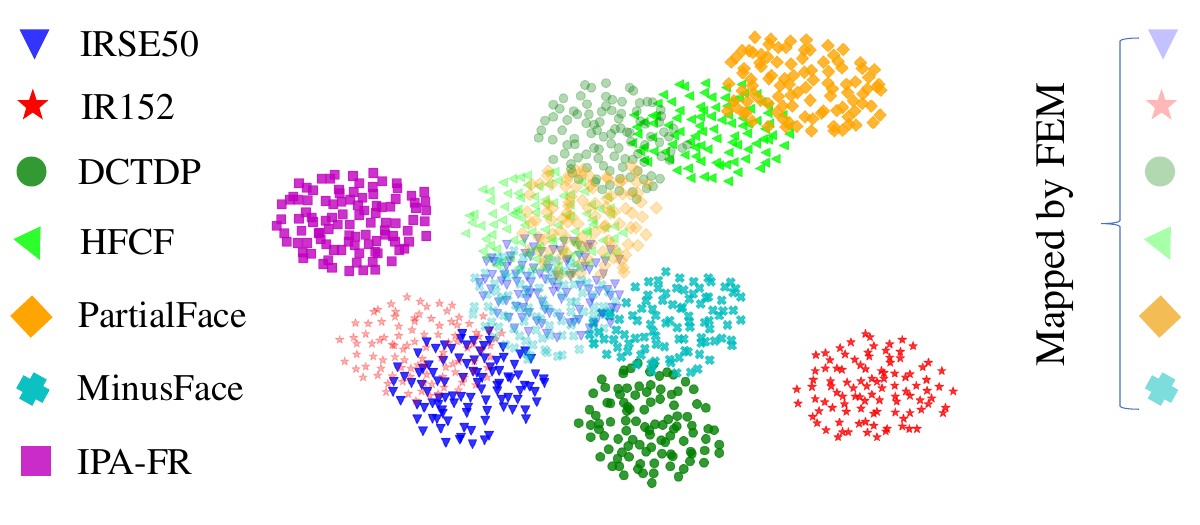}
    \caption{Face embedding distributions before and after mapped by FEMs. Visualized by UMAP \cite{mcinnes2018umap}.}
    \label{fig:embedding distribution}
\end{figure}
\begin{figure}[t!]
    \centering
    \includegraphics[scale=0.25]{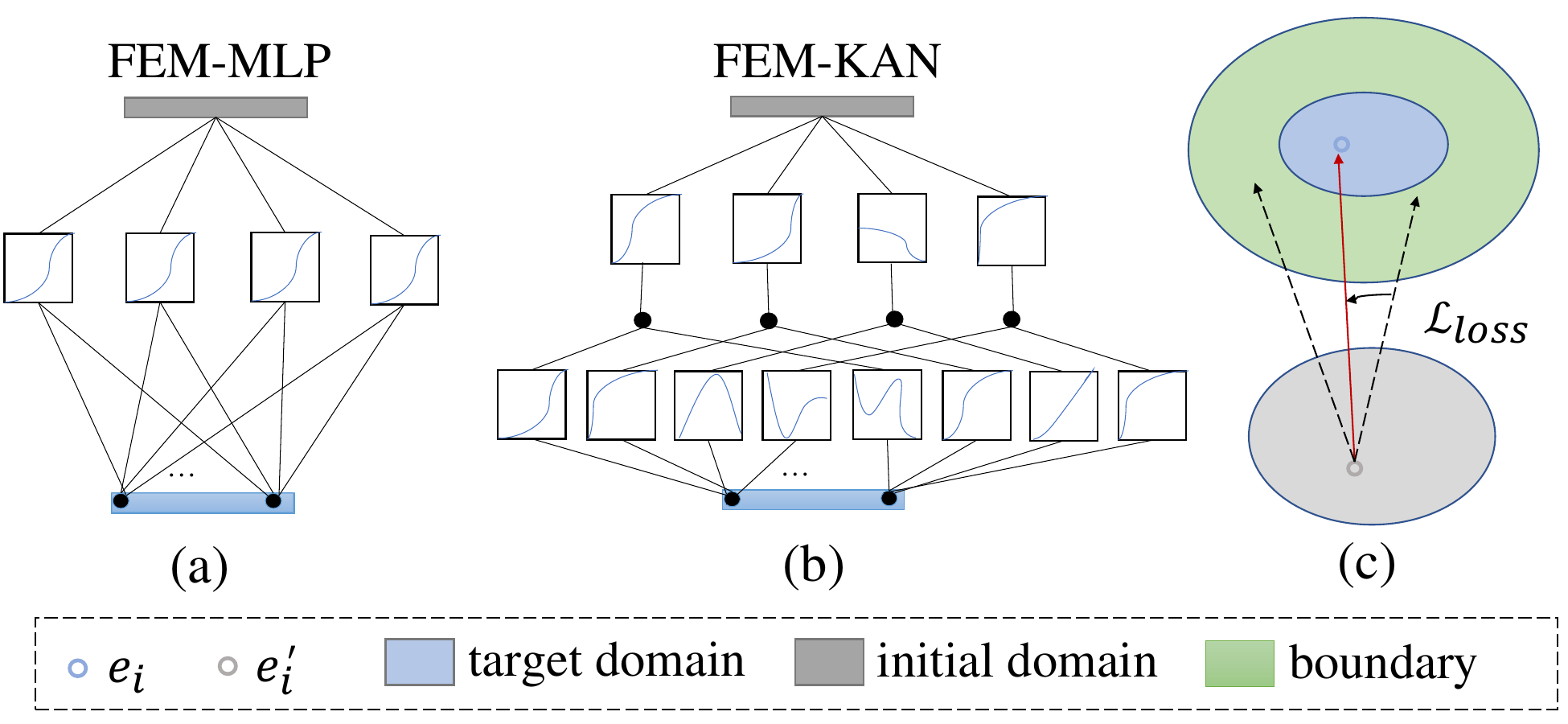}
    \caption{Two variants of FEM models and the process of embedding-to-embedding mapping. (a) FEM-MLP has fixed activation function.
    (b) FEM-KAN has learnable activation function at edges to achieve accurate non-linear mapping. (c) The direction of embedding mapping optimized by distance towards to 
    ``ground truth'' face embedding $e_i$.}
    \label{fig: FEM}
\end{figure}

Face embedding is a vector that represents facial information associated with a corresponding identity and stores soft-biometric information.
Previous work \cite{terhorst2020beyond} shows demographic information and social traits can be extracted by analyzing face embeddings of FR models. 
And prediction accuracy is associated with encoding ability of feature extractor. 
PPFR models employ different types of approaches such as transformation, noise, perturbation to protect privacy. 
Due to the nature of protection mechanism, i.e., losing information in image space or embedding space, PPFR models sacrifice a certain level of recognition ability. 
Consequently, soft-biometric information of embeddings can be diminished.
Intuitively, it becomes more difficult to extract underlying facial attributes or reconstruct facial image based on PPFR embeddings \cite{MinusFace, zhang2024validating} than normal FR embeddings.
Ideally, embeddings that are extracted from different face images of the same identity should be close and far for
those that computed from different ones. 
Existing SOTA FR and PPFR networks utilize similar structures of backbone to extract features from the face image and compute the face embedding. 
We assume there is a transformation or mapping algorithm between embeddings from the same identity that are extracted by different backbones.
Inspired from \cite{arc2face} and \cite{kan}, we propose FEM-MLP and FEM-KAN showing in Fig. \ref{fig: FEM} to
learn the mapping relation of embedding distributions from different FR and PPFR backbones. 
Then trained FEMs can map face embedding from the initial domain into the corresponding target domain of the pre-trained IPA-FaceID diffusion model
in order to generate face images, see in Fig. \ref{fig:embedding distribution}.
Depending on the effectiveness of FEMs, the mapped embedding can fall into the target domain and boundary region. 
The target domain represents mapped embedding can be used for ID-Preserving face image generation that can fool the evaluation FR systems while boundary region
indicates mapped embedding is not sufficient for ID-Preserving face image generation but human-like image generation.

For training our face reconstruction framework, considering a target FR or PPFR model $\varGamma'(.)$ and the default FR (IPA-FR) $\varGamma(.)$ model of IPA-FaceID, 
for training image dataset $ \mathcal{I}=I_i$, we can generate the embedding distribution $\mathcal{D}({e_i})$ 
as well as $\mathcal{D}'({e'_i})$ by extracting face embeddings from all face images in public or synthetic face dataset $\mathcal{I}$, where $e_i$ and $e'_i$ denote the output
face embeddings from $\varGamma(.)$ and $\varGamma'(.)$.
In order to enable target $\varGamma'(.)$ model to generate realistic target identity face images by leveraging ID-Preserving capability of the pre-trained IPA-FaceID, 
the target embedding $e'_i$ extracted from $\varGamma'(.)$ should be mapped close to the corresponding embedding $e_i$ that represents the same face identity.
Therefore, 
we should minimize the distance between $\mathcal{\hat{D}}({\hat{e}_i})$ $=$ $\mathcal{M}(\mathcal{D}'({e'_i}))$ and $\mathcal{D}({e_i})$, where $\mathcal{M}(.)$ and 
$\hat{e}_i$ denote FEM and mapped face embedding, respectively, by following reconstruction loss function (Mean Square Error):
    \begin{equation}
        \mathcal{L}_\text{MSE}(e_i, \hat{e}_i) = \frac{\sum_{i=0}^{N - 1} (e_i - \hat{e}_i)^2}{N}
    \end{equation}

\section{Experiments}
\label{sec:experi}

\subsection{Experimental Details}

Our goal is to reconstruct complete face image from embedding of target FR and PPFR models. 
The generated face images are used to fool other real-word FR systems in order to verify the reconstruction performance.  
We mainly conduct three different experiments to verify the proposed method as follows: 

\begin{itemize}
    \item For evaluating the effectiveness, we reconstruct five face images for each embedding that extracted from target FR and PPFR models and 
    inject the generated faces to the test FR models for performing transferable face verification.  
    \item For evaluating the generalization, 
    we train models on Flickr-Faces-HQ (FFHQ) \cite{FFHQ_dataset} dataset and 
    test on CelebA-HQ \cite{CelebA-HQ}, with 1000 images of never-before-seen identities. Moreover, we also investigate the impact of markup to face reconstruction on LADN dataset \cite{ladn}.
    \item For evaluating the robustness, we conduct experiments to reconstruct faces from partial embeddings and protected embeddings, e.g., PolyProtect \cite{PolyProtect}, MLP-Hash \cite{Mlp-hash} and SlerpFace \cite{slerpface}.
    Besides, we also utilize embeddings of facial images safeguarded by the facial privacy algorithm Fawkes \cite{shan2020fawkes} to reconstruct faces.
\end{itemize}

\subsection{Settings} 
\textbf{Baselines.}
We compare FEMs with \textbf{FaceTI} \shortcite{MappingNetworks} and training-free method \textbf{MAP2V} \cite{zhang2024validating}.
We train FaceTI for 20 epochs with batch size 6, using default configurations in official implementation\footnote{https://gitlab.idiap.ch/bob/bob.paper.neurips2023\_face\_ti}.
For MAP2V, we use the same setting as in official implementation\footnote{https://github.com/Beauty9882/MAP2V}. 
To showcase the effectiveness of FEMs, we also directly use pre-trained IPA-FaceID to generate face images for comparison, which is denoted as \textit{None}.  

\textbf{Datasets.} We conduct experiments on two face benchmarks, i.e., CelebA-HQ \cite{CelebA-HQ}, LADN \cite{ladn}. 

\textbf{Target models.} For target models that we aim to reconstruct face image from can be categorized into normal FR models such as \textbf{IRSE50} \cite{IR_SE}, \textbf{IR152}  \cite{IR152}
and PPFR models including \textbf{DCTDP} \cite{DCTDP}, \textbf{HFCF} \cite{HFCF}, \textbf{PartialFace} \cite{PartialFace} and \textbf{MinusFace} \cite{MinusFace}.

\textbf{Evaluation metrics.} Following \cite{shamshad2023clip2protect,sun2024diffam}, 
we employ the attack success rate (ASR) as a metric to evaluate attack efficacy of reconstructed face images from target FR and PPFR systems. 
ASR is defined as the fraction of generated faces successfully recognized by real-word FR systems. 
We generate five images for each embedding.
When determining the ASR, we establish a False Acceptance Rate (FAR) of 0.01 for each FR model.
We have selected four widely-used public FR models as real-word models for face verification to measure reconstruction performance. 
These models are ElasticFace (\textbf{EF}) \cite{elasticface}, MobileFace (\textbf{MF}) \cite{mobilefacenets}, GhostFaceNet (\textbf{GF}) \cite{ghostfacenets}, ArcFace (\textbf{AF}) \cite{IR152}. 
We use implementation from \textit{DeepFace}\footnote{https://github.com/serengil/deepface} for last two models. 

\textbf{Implementation details.}
FR \textit{buffalo\_l}\footnote{https://github.com/deepinsight/insightface} is selected as the default FR model of IPA-FaceID. 
We choose \textit{faceid\_sd15}\footnote{https://huggingface.co/h94/IP-Adapter-FaceID/tree/main} checkpoint for IPA-FaceID which takes face embedding and text as input.
In order to effectively generate face in proper angle, we fix the text prompt as ``front portrait of a person'' for all the experiments.  
As our finding, few layers are sufficient for FEMs achieving good mapping performance.
Therefore, we implement FEM-KAN with three KAN layers (follow \textit{efficient\_kan}\footnote{https://github.com/Blealtan/efficient-kan}) and FEM-MLP with three MLP layers. 
We use the GELU activation function and add 1D batch normalization to FEM-MLP. 
We train two variant FEM models with 90\% FFHQ dataset for 20 epochs.
For optimizers, we use AdamW with initial learning rate $10^{-2}$ and the exponential learning rate decay is set to 0.8.
Our experiments are conducted on a Tesla V100 GPU with 32G memory using PyTorch framework, setting the batch size
to 128 for FEMs.

\section{Results}\label{sec:results}

\subsection{Performance on Privacy Attacks}

\begin{figure}[t!]
    \centering
    \includegraphics[scale=0.37]{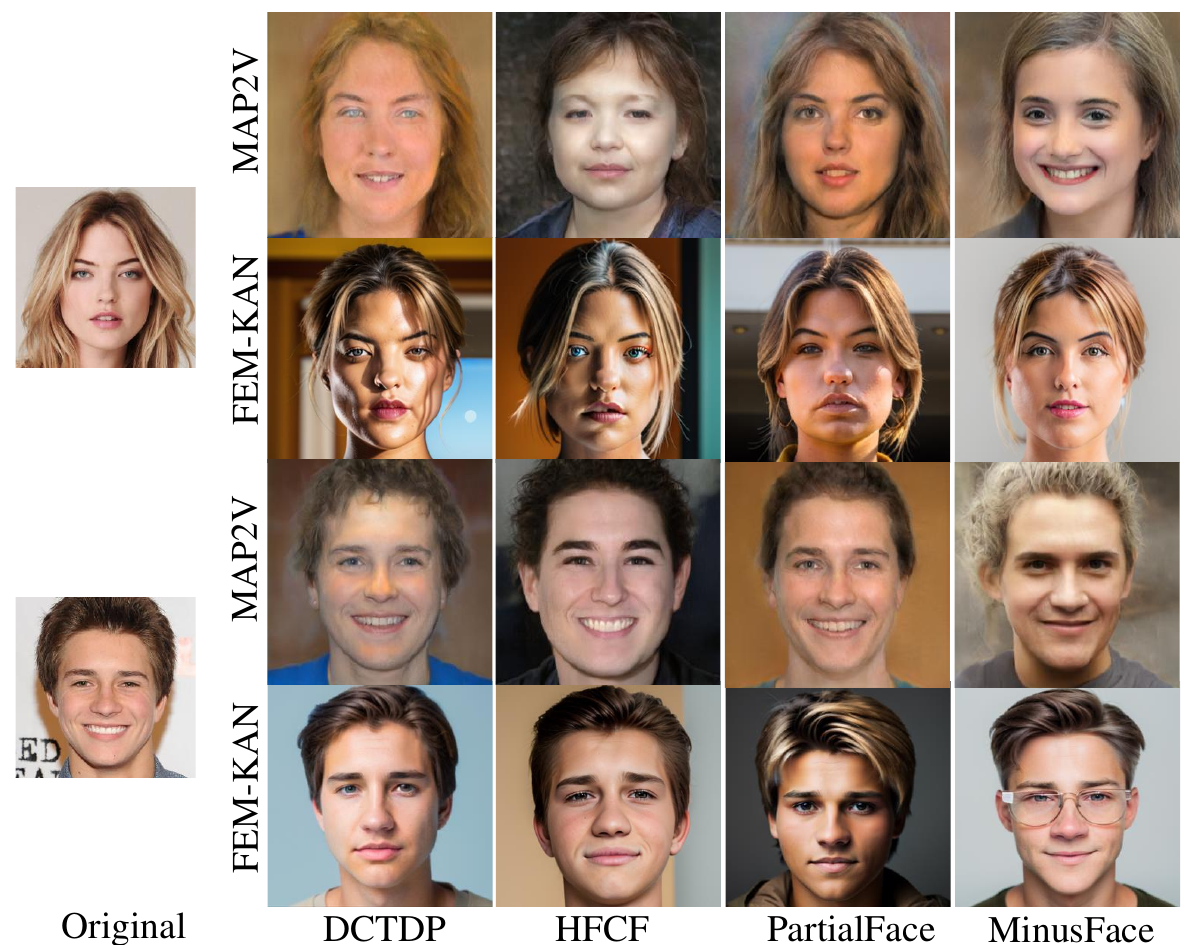}
    \caption{Visual Comparison of reconstructed faces from CelebA-HQ dataset.}
    \label{fig:PPFR reconstruction}
\end{figure}

As shown in Tab. \ref{tab:benchmark PPFR}, our proposed FEMs can effectively reconstruct face images from target FR models and outperform FaceTI and MAP2V.
Besides, FEMs also maintain high ASRs on various target PPFR models.
It shows that even images transferred into frequency domain, the corresponding face embedding still contains information
can be used for high-quality face reconstruction.
FEMs achieve higher average ASR on all PPFR models than existing SOTA method MAP2V, especially on HFCF and MinusFace.  
We exclude training PPFR models with FaceTI due to the constraints of our computational resources and the extremely long training time required by this method, 
details about resource requirements are in the ablation study. We show a visual comparison of reconstructed face images in Fig. \ref{fig:PPFR reconstruction}.
{
\setlength{\tabcolsep}{1mm}
\begin{table}[t!]
    \small
    \centering
    \begin{tabular}{ll|cccc|c}
        \hline
        \textbf{Target FR} & \textbf{Method}  & \textbf{MF}  & \textbf{EF}  & \textbf{GF} & \textbf{AF} & \textbf{Average} \\
        \hline
        \multirow{4}{*}{IRSE50} 
        & \textit{None} & 10.7 & 7.3 & 3.0 & 9.1 & 7.5\\
        & FaceTI & 93.4 & 80.8 & 49.6 & 66.8 & 72.7 \\
        & MAP2V        & 94.0 & 86.2 & 59.3  & 72.0  & 77.9  \\
        & FEM-MLP & 98.0  & 91.8 & 62.6  & 73.4 & 81.5\\
        & FEM-KAN & \cellcolor{LightCyan}\textbf{99.2} & \cellcolor{LightCyan}\textbf{93.8} & \cellcolor{LightCyan}\textbf{65.7} &\cellcolor{LightCyan}\textbf{76.1} & \cellcolor{LightCyan}\textbf{83.7} \\
        \hline
        \multirow{4}{*}{IR152} 
        & \textit{None} & 9.2 & 5.8 & 7.6 & 2.3 & 6.2\\
        & FaceTI & 85.2  & 74.0 & 43.4 & 61.2 & 66.0\\
        & MAP2V        & 92.3  & 82.5 & 53.0  & 67.2  & 73.8  \\
        & FEM-MLP      & 94.1  & 78.1 & 48.6 & 64.6 & 71.4 \\
        & FEM-KAN & \cellcolor{LightCyan}\textbf{95.0}  & \cellcolor{LightCyan}\textbf{85.2} & \cellcolor{LightCyan}\textbf{58.4} & \cellcolor{LightCyan}\textbf{70.3} & \cellcolor{LightCyan} \textbf{77.2} \\
        \hline
        \textbf{Target PPFR} &   &  &  &  &  &  \\
        \hline
        \multirow{4}{*}{DCTDP} & \textit{None} & 6.8 & 5.4 & 2.9 & 7.1 & 5.6 \\
        & MAP2V   & 94.0 & 85.5 & 59.4  & 74.3  & 78.3 \\
        & FEM-MLP & 97.3 & \cellcolor{LightCyan}\textbf{91.8} & 67.9 & 77.7 & 83.7\\
        & FEM-KAN & \cellcolor{LightCyan}\textbf{98.5} & \cellcolor{LightCyan}\textbf{91.8} & \cellcolor{LightCyan}\textbf{68.4} & \cellcolor{LightCyan}\textbf{78.8} & \cellcolor{LightCyan}\textbf{84.4} \\
        \hline
        \multirow{3}{*}{HFCF} & \textit{None} & 7.5 & 4.2 & 5.1 & 11.4 & 7.1\\
        & MAP2V   & 76.3  & 15.4  & 5.3  & 14.8 & 28.0 \\
        & FEM-MLP & 97.0 & 89.2 & 61.8 & 74.2 & 80.6\\
        & FEM-KAN & \cellcolor{LightCyan}\textbf{98.3} & \cellcolor{LightCyan}\textbf{90.7} & \cellcolor{LightCyan}\textbf{66.5} & \cellcolor{LightCyan}\textbf{76.9} & \cellcolor{LightCyan}\textbf{83.1}\\
        \hline
        \multirow{3}{*}{PartialFace} & \textit{None} & 7.9   &5.8   &1.8   &6.6   & 5.5  \\
        & MAP2V   & 93.1 & 84.3 & 60.0  & 71.4  & 77.2 \\
        & FEM-MLP & 98.7  & 88.5 & 61.5 & 71.6 & 80.1  \\
        & FEM-KAN & \cellcolor{LightCyan}\textbf{99.5} & \cellcolor{LightCyan}\textbf{90.5} & \cellcolor{LightCyan}\textbf{68.0} & \cellcolor{LightCyan}\textbf{77.6} & \cellcolor{LightCyan}\textbf{83.9}\\
        \hline
        \multirow{3}{*}{MinusFace} & \textit{None} &3.5  &5.1  &2.1  &10.3  &5.3  \\
        & MAP2V   &68.0  &4.8   & 2.3  & 5.6  & 20.2 \\
        & FEM-MLP &94.4  &68.5  &\cellcolor{LightCyan}\textbf{45.9}  &\cellcolor{LightCyan}\textbf{59.4}  &  67.1\\
        & FEM-KAN & \cellcolor{LightCyan}\textbf{96.5} & \cellcolor{LightCyan}\textbf{71.3} & 44.5 & 58.1 & \cellcolor{LightCyan}67.6\\
        \hline
    \end{tabular}
    \caption{Evaluations of ASR for black-box attacks to \textbf{FR} and \textbf{PPFR} models on CelebA-HQ dataset.}
    \label{tab:benchmark PPFR}
\end{table}
}

\subsection{Makeup Reconstruction From Face Embeddings}
Previous work shows \cite{chen2017spoofing} makeup can be used as markup presentation attacks, e.g., for impersonation.
For instance, attacker can apply specific markup such that the face of attacker looks similar to that of a target subject.
However, there is no existing work studying how markup will affect the face reconstruction.
Therefore, apart from reconstructing natural-looking face images from face embeddings,
we also consider a more challenging scenario to evaluate whether the proposed method can also reconstruct markups applied to face images that were used for extracting face embeddings.
As shown in Tab. \ref{tab: attack LADN dataset}, we demonstrate the impact of markup to face reconstruction.
Markup negatively affects the ASR of FaceTI resulting in significant accuracy decreasing, i.e., 20.5\% on GF and 26.6\% on AF, respectively.
Moreover, FaceTI only achieves 56.4\% average ASR on markup dataset.
MAP2V also suffers a noticeable average ASR drop around 8.8\%.
In contrast, our proposed FEMs are more robust to markup and maintain high performance in reconstructing face images with markup, e.g., 85.1\% ASR in average.
{
\setlength{\tabcolsep}{1mm}
\begin{table}[t!]
    \small
    \centering
    \begin{tabular}{l l | c c c c |c}
        \hline
        \centering
        \textbf{Dataset} & \textbf{Method}  & \textbf{MF} & \textbf{EF} & \textbf{GF} & \textbf{AF} & \textbf{Average} \\
        \hline
        \multirow{3}{*}{\textbf{LADN-NM}} 
        & FaceTI & 98.5  &75.4  &50.6  &73.4  & 74.5 \\
        & MAP2V        & 98.2  &93.1  &66.8  &79.0  & 84.3 \\
        & FEM-MLP & \cellcolor{LightCyan}\textbf{100.0}  &94.0  &72.5  &83.5  & 87.5 \\
        & FEM-KAN & \cellcolor{LightCyan}\textbf{100.0} & \cellcolor{LightCyan}\textbf{94.9} & \cellcolor{LightCyan}\textbf{78.7} & \cellcolor{LightCyan}\textbf{92.5} & \cellcolor{LightCyan}\textbf{91.5}\\
        \hline
        \multirow{3}{*}{\textbf{LADN-M}} 
        & FaceTI  &91.0  &57.7  &30.1  &46.8  & 56.4 \\
        & MAP2V        & 97.2  &85.6  &51.3  &67.9  & 75.5 \\
        & FEM-MLP & 99.4  & 87.9 & 63.7 & 73.5 & 81.1  \\
        & FEM-KAN & \cellcolor{LightCyan} \textbf{100.0} & \cellcolor{LightCyan}\textbf{94.9} & \cellcolor{LightCyan}\textbf{69.3} & \cellcolor{LightCyan}\textbf{76.3} & \cellcolor{LightCyan}\textbf{85.1}\\
        \hline
    \end{tabular}
    \caption{ASR performance on reconstructing face images with markup. \textbf{LADN-NM} and \textbf{LADN-M} denote non-markup and markup datasets.}
    \label{tab: attack LADN dataset}
  \end{table}
}
\subsection{Face Reconstruction from Partial Leaked Embeddings}
Previous experiments are based on the assumption that the adversary can gain access to the complete face embeddings. 
Nevertheless, in some real-world scenarios, a complete face embedding is difficult to acquire, but rather to access a portion of the embedding.
For example, face embeddings of the FR system are split and stored on different servers for data protection like the situation considered in \cite{partial_face_reconstruction}.
We assume the adversary already trained FEMs on complete embeddings of the target FR or PPFR model. 
To further test non-linear mapping ability and face reconstruction, we only use partial leaked embeddings (e.g., discarding the half values in an embedding vector in the case of 50\% leakage) as input to trained FEMs.
To match the input shape of FEMs, we append zeros to the end of each leaked embedding vector.
\begin{figure}[t!]
    \centering
    \includegraphics[scale=0.245]{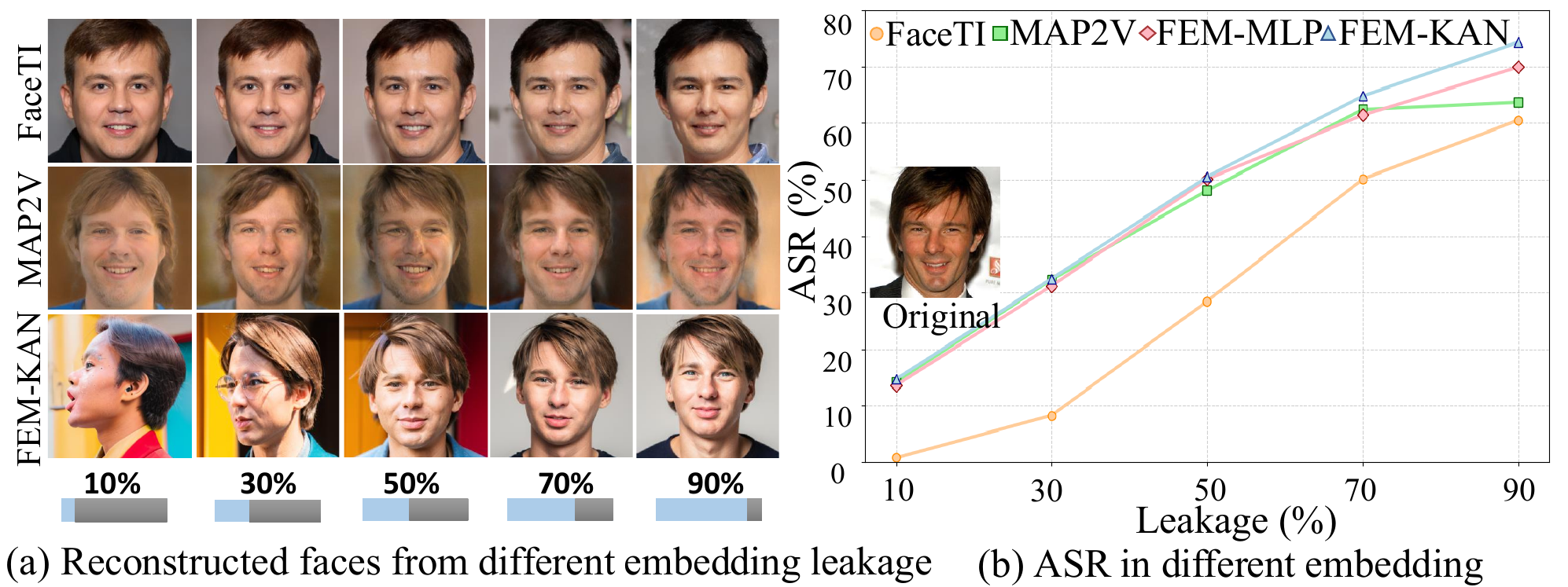}
    \caption{(a) Reconstructed face comparison with different embedding leakages. (b) ASR performance evaluated by ArcFace. IRSE50 is target model.}
    \label{fig: reconstructed faces from partial embedding}
\end{figure}
In Fig. \ref{fig: reconstructed faces from partial embedding} (a),
the reconstructed face images can reveal privacy-sensitive information about the underlying user, such as gender, race, etc.
FaceTI and MAP2V can reconstruct general attributes (gender, age and facial expression) in the case of high leakage, e.g., 70\% to 90\% while our method is able to reconstruct more detailed attributes similar to the original identity.
Under 50\% leakage scenario, FaceTI start to generate unified and normalized faces for embeddings belonging to different identities.
In contrast, FEM can still reconstruct the general facial structures of identity.
However, the generated face tends to have noticeable artifacts when leakage is lower than 30\%. 
Fig. \ref{fig: reconstructed faces from partial embedding} (b) reports ASR to evaluate the incomplete leaked embedding mapping ability of FEMs.
With an increased percentage of embedding leakage, the number of generated face images that can fool the test FR is reduced.
FaceTI suffers the significant impact with decreased embedding leakage, i.e., 22\% ASR reduction when 50\% leakage compared with MAP2V and FEMs.
For the extreme low leakage below 30\%, FEMs still achieve relatively high ASR (32.5\% for FEM-KAN). 
Therefore, our method is more robust than GAN-based method and able to reconstruct facial information from incomplete embeddings.
\subsection{Face Reconstruction from Protected Embeddings}
{
\setlength{\tabcolsep}{1mm}
\begin{table}[t!]
    \small
    \centering
    \begin{tabular}{ l l| c c c c c}
        \hline
        \textbf{Protection} & \textbf{Method} & \textbf{MF} & \textbf{EF} & \textbf{GF} & \textbf{AF} \\
        \hline
        \multirow{3}{*}{PolyProtect} 
        & FaceTI  & 45.6  & 3.3   &0.8  & 4.1 \\ 
        & MAP2V   & 28.6  & 4.4   &3.6  & 4.3 \\ 
        & FEM-MLP & 48.3  & 4.7   & \cellcolor{LightCyan}\textbf{6.8} & \cellcolor{LightCyan}\textbf{15.6}  \\
        & FEM-KAN & \cellcolor{LightCyan}\textbf{50.3}   & \cellcolor{LightCyan}\textbf{7.1}    &5.6  & 15.4 \\
        \hline
        \multirow{4}{*}{MLP-Hash} 
        & FaceTI & 66.8  & 0.7   &0.1  & 0.4 \\    
        & MAP2V   & 48.1  & 0.6   &0.3  & 1.5 \\                              
        & FEM-MLP & 80.4  & 53.0   & 51.9  & 64.5   \\
        & FEM-KAN & \cellcolor{LightCyan}\textbf{82.1}  & \cellcolor{LightCyan}\textbf{54.7}   & \cellcolor{LightCyan}\textbf{56.5}  & \cellcolor{LightCyan}\textbf{71.6} \\   
        \hline
        \multirow{4}{*}{SlerpFace} 
        & FaceTI  & 24.6  & 1.2   & 0.3  & 0.1  \\    
        & MAP2V   & 11.4  & 0.0  & 0.1 & 0.1 \\                              
        & FEM-MLP & 78.6  & 7.1   & 4.5  & 14.1   \\
        & FEM-KAN & \cellcolor{LightCyan}\textbf{79.4}  & \cellcolor{LightCyan}\textbf{9.3}   & \cellcolor{LightCyan}\textbf{7.8}  & \cellcolor{LightCyan}\textbf{15.4} \\    
        \hline
    \end{tabular}
    \caption{ASR performance on protected face embeddings.}
    \label{tab: ASR on protected embedding}
\end{table}
}

Considering more strict access to the original embeddings that were directly computed by the feature extractor of PPFR models, 
we evaluate FEMs on face embeddings that being protected by particular embedding protection algorithms such as \textbf{PolyProtect} \cite{PolyProtect}, \textbf{MLP-Hash} \cite{Mlp-hash} and \textbf{SlerpFace} \cite{slerpface}.
We train FEMs directly on protected face embeddings derived from these protection algorithms based on PPFR HFCF model.
For PolyProtect, we generate the user-specific pair for each identity in the testing dataset.
After mapping the original face embedding from PPFR model, the protected embedding from PolyProtect has reduced dimension, 508 in our setting. During training, we append other four zeros to the end of the protected embedding to maintain the length of vector.
For MLP-Hash method, we set one-hidden layer with 512 neurons and fix the seed for all identities. As for SlerpFace, we use the default settings in original implementation.
Tab. \ref{tab: ASR on protected embedding} reports the ASR on the protected embeddings. 
It is worth noticing that FEMs still can achieve high face reconstruction against MLP-Hash and have comparable ASR with the ones on unprotected embeddings in Tab. \ref{tab:benchmark PPFR}.
Moreover, FEMs have the highest ASRs against three protection algorithms than FaceTI and MAP2V.
{
\setlength{\tabcolsep}{1mm}
\begin{table}[t!]
    \small
    \centering
    \begin{tabular}{ll| cccc}
        \hline
        \textbf{Protection} & \textbf{Method} & \textbf{MF} & \textbf{EF} & \textbf{GF} & \textbf{AF} \\
        \hline
        \multirow{4}{*}{Fawkes} 
        & FaceTI & 85.0  & 16.3   &6.3  & 13.0 \\ 
        & MAP2V        & 96.3  & 44.9   & 17.4  & 25.7 \\ 
        & FEM-MLP & \cellcolor{LightCyan}\textbf{97.0}   & 42.0    & 16.7   & 24.2  \\
        & FEM-KAN & \cellcolor{LightCyan}\textbf{97.0}   & \cellcolor{LightCyan}\textbf{47.1}    &\cellcolor{LightCyan}\textbf{19.0}  & \cellcolor{LightCyan}\textbf{27.9} \\
        \hline
    \end{tabular}
    \caption{ASR performance on protected face images.}
    \label{tab: ASR on protected facial image}
\end{table}
}
\subsection{Face Reconstruction from Protected Facial Images}
Unauthorized FR systems identifying and recognizing 
online photos pose a threat to personal security and
privacy in the digital world.
Most of existing countermeasures for protecting privacy of facial images are based on 
introducing perturbations \cite{madry2018towards,shan2020fawkes} or adding markups \cite{sun2024diffam} in pixel level to misguide unauthorized FR systems.
Intuitively, it is more difficult to reconstruct facial image from leaked embedding of protected image.
In this section, we focus on evaluating FEMs against SOTA facial protection algorithm \textbf{Fawkes} \cite{shan2020fawkes} on IRSE50.
It is a \textit{black-box} cloaking algorithm to add imperceptible perturbations. 
Considering computational cost and strength of privacy protection, we apply Fawkes to CelebA-HQ using mid level perturbations to generate cloaked version of dataset. 
As depicted in Tab. \ref{tab: ASR on protected facial image}, ASRs are considerly reduced across on different public FR systems compared 
with performance (see in Tab. \ref{tab:benchmark PPFR}) on reconstructing face images from embeddings of corresponding clean dataset.
It is worth to notice that reconstructed face images still maintain high ASRs to access MobileFace, which can threaten to other lightweight FR systems that designed for running face recognition algorithms with limited hardware resources, e.g., phone and tablet.
Therefore, the more powerful face image protection algorithms should be developed to further reduce the risk of face reconstruction attack without introducing noticeable artifacts.
{
\setlength{\tabcolsep}{1mm}
\begin{table}[t!]
    \small
    \centering
    \begin{tabular}{l|ccc}\hline 
                        & \textbf{Training Time}  & \textbf{Memory} & \textbf{Inference Time} \\\hline
          FaceTI       & 51 hrs   & 25383 MiB                 & 3.2s      \\
          MAP2V         & 0        & 32133 MiB                 & 111s      \\
          FEMs          & 3  hrs   & 4325  MiB                 & 2.6s     \\\hline
  \end{tabular}
  \caption{Training resource requirements when training (single epoch) on 90\% FFHQ dataset and inference time on reconstruct a single image.}\label{tab: efficient} 
\end{table}
}
\subsection{Ablation Study}\label{subsec:ablation}
\textbf{Training and inference efficacy.} 
To produce human-like images using pre-trained StyleGAN3 model, FaceTI requires training an additional critic network (Wasserstein GAN \cite{arjovsky2017wasserstein}) together with the mapping network simultaneously, which demands high computational cost and is resource-intensive. 
In contrast, FEMs require \textbf{17$\times$} less training time and \textbf{5.8$\times$} less GPU memory compared with FaceTI, see in Tab. \ref{tab: efficient}.
We also measure inference time that reconstructing a complete face from an embedding. 
MAP2V takes extreme long time for single image generation.
Our method can reconstruct one image within 3 seconds, which is around \textbf{42$\times$} faster than MAP2V.

{
\setlength{\tabcolsep}{1mm}
\begin{table}[t!]
    \small
    \centering
    \begin{tabular}{l|ccccc}
        \hline
        \centering
       \textbf{Method}  & \textbf{MF}  & \textbf{EF}  & \textbf{GF} & \textbf{AF} & \textbf{Average} \\
        \hline
        \centering
        FaceTI & 66.4 & 54.4 & 43.6 & \cellcolor{LightCyan}\textbf{69.8} & 58.6 \\
        MAP2V & 86.8 & 66.2 & 52.8 & \cellcolor{LightCyan}\textbf{54.2}  & \cellcolor{LightCyan}\textbf{65.0}  \\
        FEM-MLP & \cellcolor{LightCyan}\textbf{89.0}  & 67.2 & 38.4  & 46.6 & 60.3\\
        FEM-KAN & 88.0  & \cellcolor{LightCyan}\textbf{69.0} & 45.4 & 50.8 & 63.3 \\
        \hline
    \end{tabular}
    \caption{Evaluations of ASR for black-box attacks to IRSE50 on LFW dataset.}
    \label{tab:benchmark_lfw_test}
\end{table}
}

\textbf{Reconstruction from low-resolution facial images.}
For scenario that leaked embedding is from low-resolution face,
we randomly select 500 images from LFW \cite{huang2008labeled}, with image size in 112 $\times$ 112.
As shown in Tab. \ref{tab:benchmark_lfw_test}, 
we observe dropped ASRs compared with the performance on high-resolution CelebA-HQ images. 
\begin{figure}[t!]
    \centering
    \includegraphics[scale=0.249]{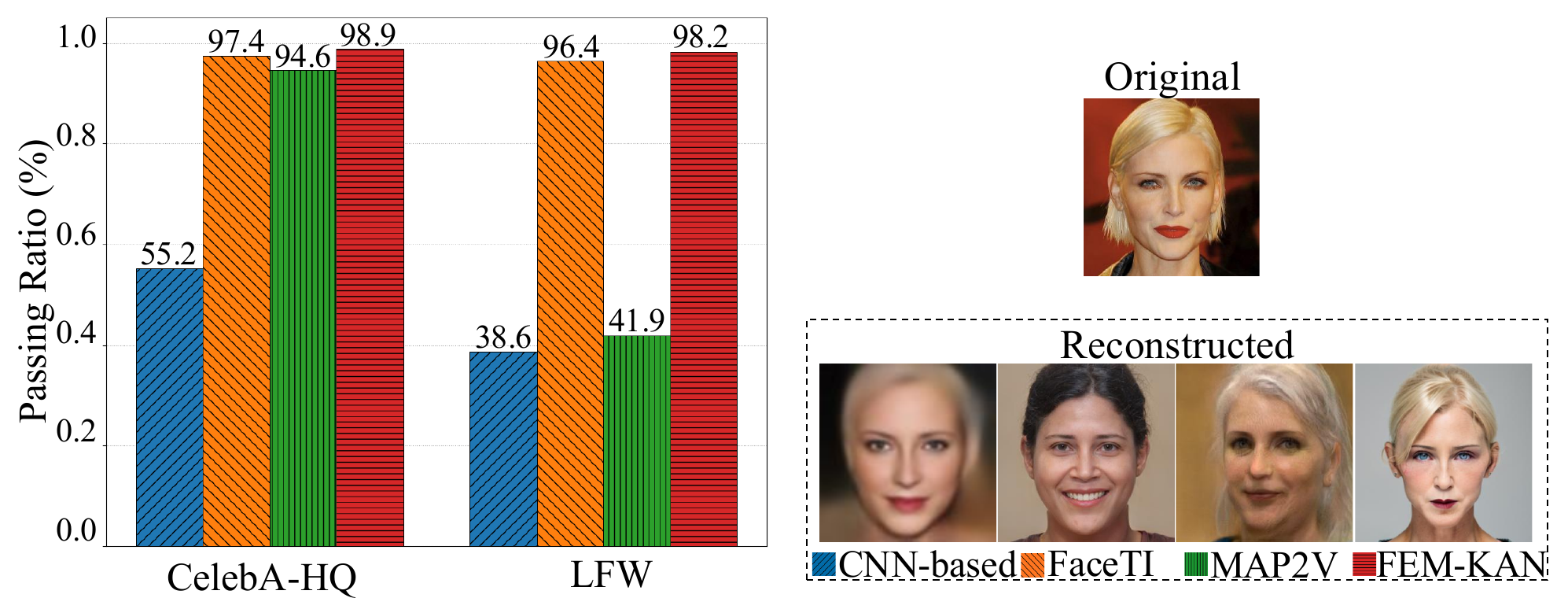}
    \caption{The percentage of reconstructed images passing face anti-spoofing system.}
    \label{fig: real ratio}
\end{figure}
\textbf{Face anti-spoofing.}
Face Anti-Spoofing (FAS) is a critical security measure designed to detect and prevent attempts to deceive FR systems. 
By assuming real-word FR systems incorporate FAS to reject potential fake face images before recognition, 
we conduct experiment with FASNet \cite{lucena2017transfer} to evaluate the attacking effectiveness by passing ratio, which is defined as the fraction reconstructed faces 
bypassing the FAS system. In Fig. \ref{fig: real ratio}, we also add CNN-based method \cite{shahreza2023blackbox} to demonstrate why it is not practical for real-word attack. 
Since reconstructed face images from CNN-based method are not realistic (e.g., blurry, noisy artifacts) and have low resolutions,
these images are easily rejected by FAS system. MAP2V faces the same issue when reconstructing from LFW dataset. 

\section{Conclusion}
\label{sec:conclusion}

In this paper, we propose a new framework called FEM to reconstruct high-resolution, realistic face images from face embeddings in both FR and PPFR systems.
FEM can project arbitrary face embeddings for face reconstruction by leveraging a pre-trained face image generation network, IPA-FaceID.
We measure the effectiveness in different scenarios, including \textit{black-box} embedding-to-face attacks, out-of-distribution generalization, 
reconstructing from partial leaked embeddings, protected embeddings and protected facial images.
Extensive experiments demonstrate that FEM outperforms SOTAs in both FR and PPFR scenarios. 
Moreover, FEM framework can be served as an evaluation tool to verify the robustness of existing PPFRs and privacy protection algorithms.

\bibliography{aaai2026}

\end{document}